\documentclass{article}

\usepackage[preprint]{neurips_2025}

\usepackage{multirow}
\usepackage{amsmath}

\usepackage[utf8]{inputenc} 
\usepackage[T1]{fontenc}    
\usepackage{hyperref}       
\usepackage{url}            
\usepackage{booktabs}       
\usepackage{amsfonts}       
\usepackage{nicefrac}       
\usepackage{microtype}      
\usepackage{xcolor}         
\usepackage{graphicx}
\usepackage{subcaption}
\newcommand{\name}{EpiLLM}

\title{EpiLLM: Unlocking the Potential of Large Language Models in Epidemic Forecasting}

\author{%
  Chenghua Gong \footnotemark[1]\\
  \small{University of Science and Technology of China}\\
  \small{Hefei, China} \\
  \texttt{gongchenghua888@gmail.com} \\
  \And
  Rui Sun \thanks{Co-first authors with equal contribution.}\\
  \small{University of Science and Technology of China} \\
  \small{Hefei, China} \\
  \texttt{gm1984226519@gmail.com} \\
  \And
  Yuhao Zheng \\
  \small{University of Science and Technology of China} \\
  \small{Hefei, China} \\
  \texttt{yuhaozheng@mail.ustc.edu.cn} \\
  \And
  Juyuan Zhang \\
  \small{University of Science and Technology of China} \\
  \small{Hefei, China} \\
  \texttt{zhangjuyuan2020@mail.ustc.edu.cn} \\
  \And
  Tianjun Gu \\
  \small{East China Normal University} \\
  \small{Shanghai, China} \\
  \texttt{51275901043@stu.ecnu.edu.cn} \\
  \And
  Liming Pan \\
  \small{University of Science and Technology of China} \\
  \small{Hefei, China} \\
  \texttt{pan\_liming@ustc.edu.cn} \\
  \And
  Linyuan L\"u \thanks{Corresponding author.}\\
  \small{University of Science and Technology of China} \\
  \small{Hefei, China} \\
  \texttt{linyuan.lv@ustc.edu.cn} \\
}

\begin{document}

\maketitle

\begin{abstract}
Advanced epidemic forecasting is critical for enabling precision containment strategies, highlighting its strategic importance for public health security.
While recent advances in Large Language Models (LLMs) have demonstrated effectiveness as foundation models for domain-specific tasks, their potential for epidemic forecasting remains largely unexplored. 
In this paper, we introduce \name, a novel LLM-based framework tailored for spatio-temporal epidemic forecasting. 
Considering the key factors in real-world epidemic transmission: infection cases and human mobility, we introduce a dual-branch architecture to achieve fine-grained token-level alignment between such complex epidemic patterns and language tokens for LLM adaptation.
To unleash the multi-step forecasting and generalization potential of LLM architectures, we propose an autoregressive modeling paradigm that reformulates the epidemic forecasting task into next-token prediction. 
To further enhance LLM perception of epidemics, we introduce spatio-temporal prompt learning techniques, which strengthen forecasting capabilities from a data-driven perspective.
Extensive experiments show that \name~ significantly outperforms existing baselines on real-world COVID-19 datasets and exhibits scaling behavior characteristic of LLMs. 
Code is available at: \href{https://anonymous.4open.science/r/EpiLLM-73C6}{https://anonymous.4open.science/r/EpiLLM-73C6}.
\end{abstract}

\section{Introduction}
Contagious epidemic outbreaks—most notably the COVID-19 pandemic~\cite{ciotti2020covid}—have emerged as some of the most significant global health emergencies in recent decades.
According to surveillance data from the World Health Organization (WHO)\footnote{Source: https://covid19.who.int/ as of May 15, 2025}, COVID-19 has resulted in approximately 778 million confirmed cases and over 7 million deaths worldwide. This has placed unprecedented strain on healthcare systems and underscored the urgent need for optimized resource allocation and public health interventions.
Consequently, accurate and timely modeling and forecasting of epidemics are critical for understanding transmission dynamics and enabling effective containment strategies.

In the wake of the COVID-19 pandemic, a wide range of epidemic modeling approaches have been developed to forecast transmission dynamics and support public health interventions~\cite{xiang2021covid}.
Early efforts typically fall into two categories: mechanistic models and statistical models. Mechanistic models aim to mathematically characterize disease transmission based on biological processes (e.g., compartmental models such as SIR/SEIR~\cite{poonia2022enhanced,wei2020fitting,hendy2021mathematical}), while statistical models focus on capturing patterns in observed data to forecast future trends~\cite{mahmud2020bangladesh,kufel2020arima}.
Despite their widespread use, both paradigms often depend on idealized assumptions or handcrafted features informed by domain expertise, limiting their adaptability and robustness in complex, real-world epidemic scenarios~\cite{roda2020difficult}.

Given the limitations of mechanistic and statistical models, recent research has increasingly focused on machine learning and deep learning approaches for epidemic forecasting~\cite{wu2018deep,wang2020prediction}.
Traditional machine learning methods such as Linear Regression~\cite{kaur2022forecasting}, Random Forests~\cite{galasso2022random}, and XGBoost~\cite{fang2022application} have been applied to predict epidemic trends with varying degrees of success. 
With the advancement of deep learning, time-series models like LSTM and other RNN variants have been adopted to capture temporal dependencies in epidemic data~\cite{chimmula2020time}.
However, these models fail to integrate spatial relations, limiting their ability to fully model the spatio-temporal nature of epidemics~\cite{panagopoulos2021transfer}. 
To this end, spatio-temporal graph neural networks~\cite{sahili2023spatio} have emerged as a promising direction, enabling the modeling of complex spatial interactions such as geographic proximity and human mobility—alongside temporal dynamics~\cite{yu2023spatio}. 
It is evident that cutting-edge deep learning techniques are continuously advancing the development of epidemic forecasting~\cite{liu2024review}.

Recent advance reveals the disruptive capacity of LLMs as foundation models across diverse fields, including financial forecasting~\cite{zhao2024revolutionizing}, cascade modeling~\cite{zheng2025autocas} and traffic accident prediction~\cite{de2023llm}.
The central idea of repurposing LLMs lies in that both natural language and temporal measurements in typical systems  share the common patterns of sequence data~\cite{liu2024autotimes}.
Existing work has begun to explore adopting LLMs as general time-series predictors~\cite{jin2023time,liu2024autotimes}, and epidemic forecasting can be abstracted as an even more complex time-series forecasting problem~\cite{panagopoulos2021transfer}, so a natural idea emerges: Can we adopt the powerful LLMs to enhance epidemic forecasting?

Epidemic forecasting is influenced by a range of complex factors, including population immunity, geospatial interactions, pathogen traits, and more~\cite{desai2019real}.
Recent work such as PandemicLLM~\cite{du2024advancing} frames this task as a complex reasoning problem and tackles it via LLM fine-tuning. 
However, fine-tuning LLMs for domain-specific forecasting is often both cost-prohibitive and technically demanding, particularly in areas lacking domain expertise~\cite{zhang2024scaling}. 
To circumvent these constraints, we focus on leveraging more readily available spatio-temporal epidemic data, such as daily infection cases and human mobility records, as input signals. 
Since recent studies have empirically demonstrated LLM potential for temporal modeling~\cite{liu2024autotimes,zheng2025autocas}, modeling epidemics solely through temporal dynamics is inherently limited~\cite{panagopoulos2021transfer,liu2024review,hy2022temporal}. 
This leads us to a key technical challenge: How can spatio-temporal epidemic data be effectively integrated within the LLMs for futher forecasting?

Prior research indicates that the strong generalization capabilities of LLMs stem primarily from autoregressive modeling~\cite{wang2022language}.
The autoregressive paradigm of predicting the next token based on a sequence of previous tokens aligns naturally with language generation and remains the dominant training strategy for LLMs. 
To extend the power of LLMs to specific domains, foundation models in fields such as vision~\cite{rajasegaran2025empirical} and time-series~\cite{liu2024autotimes} have begun reformulating their tasks as next-token prediction to align with the LLM architecture. 
In the context of epidemic forecasting, disease progression is inherently dependent on historical states~\cite{chimmula2020time}, making autoregressive modeling a natural and viable approach.
This raises a second key technical challenge: How to reformulate the epidemic forecasting task into next-token prediction?

While LLMs exhibit remarkable generalization capabilities, adapting them to domain-specific forecasting tasks can still result in suboptimal performance due to inadequate task alignment~\cite{zheng2025autocas}.
Recent studies have explored prompt learning as a means to bridge this gap~\cite{liu2023pre}, introducing textual prompts to guide LLMs toward better task integration~\cite{liu2024autotimes,zheng2025autocas}. 
However, the effectiveness and interpretability of purely textual prompts remain under debate, particularly for temporal, structured, non-linguistic data~\cite{sun2023graph,ye2024survey}. 
To fully unlock the potential of LLMs for epidemic forecasting, it is essential to account for the unique characteristics of spatio-temporal epidemic data, which are structured, dynamic, and multi-dimensional~\cite{hy2022temporal}. 
This leads to the final technical challenge: How to design the prompt learning strategy to effectively guide LLMs in epidemic forecasting task?

In this paper, we repurpose LLMs as epidemic forecasters, and introduce a novel framework named \name.
We conduct epidemic forecasting based on spatio-temporal data, focusing on two key epidemic indicators: infection case and human mobility.
Technically, we first establish a dual-branch module to capture spatio-temporal patterns for token-level modality alignment with LLM.
To further unleash the potential of LLM, we reformulate the epidemic forecasting task into next token prediction via autoregressive modeling. 
Inspired by prompt learning, we introduce spatio-temporal prompting techniques to facilitate the deeper integration of LLMs into epidemic forecasting. 
Our contributions are summarized as follows:

\begin{itemize}
\item
We innovatively integrate spatio-temporal epidemic data with the LLM architecture and introduce an LLM-based framework named \name~for epidemic forecasting.
To our best knowledge, this paper is one of the pioneering attempts to repurpose LLMs as foundation models for spatio-temporal epidemic modeling.

\item 
We reformulate epidemic forecasting into next token prediction via autoregressive modeling paradigm, and further introduce spatio-temporal prompting techniques to advance epidemic forecasting to unleash the potential of LLM architecture. 

\item 
We conduct extensive experiments on real-world COVID-19 datasets to evaluate \name.
Experimental results show that \name~significantly outperforms existing state-of-the-art competitors
in epidemic forecasting and exhibits scaling
behavior empowered by LLMs.
\end{itemize}

\section{Related Work}
\subsection{Epidemic Modeling}
Epidemic modeling plays a role in public health security. Existing methods can be broadly categorized into three types: mechanistic \& statistical models, machine learning models and deep learning models.

\paragraph{Mechanistic \& statistical models} 
Early-stage studies focuses on mechanistic and statistical models for epidemic modeling.
Mechanistic models integrate biological priors with mathematical equations to characterize epidemics under idealized conditions, with SIR and its variants being representative examples~\cite{poonia2022enhanced,wei2020fitting,hendy2021mathematical}.
Statistical models identify latent patterns via statistical characteristics of historical data to forecast future trends, with models like PROPHET~\cite{mahmud2020bangladesh} and ARIMA~\cite{kufel2020arima} being widely applied.

\paragraph{Machine learning models}
The advancements in machine learning have led to the applications of more sophisticated modes to epidemic modeling~\cite{wang2020prediction}.
Canonical methods such as Linear Regression~\cite{kaur2022forecasting}, Gaussian Process Regression~\cite{ketu2021enhanced}, Random Forest~\cite{galasso2022random}, and XGBoost~\cite{fang2022application} remain active in epidemic modeling due to their computational efficiency and rapid response.

\paragraph{Deep learning models}
Given the intricate interplay of real-world factors, deep learning has been introduced to boost data-driven epidemic modeling~\cite{wu2018deep}.
Fan et al.~\cite{fan2023exploring} examine the influence of spatial structure (e.g., geographical information, model-generated gravity) in epidemic modeling and introduce graph neural networks (GNNs) to capture these patterns.
Panagopoulos et al.~\cite{panagopoulos2021transfer} identifies human mobility as the pivotal determinant and integrates GNNs and LSTM to model spatio-temporal dynamics of epidemics within mobility networks.
Further, Hy et al.~\cite{hy2022temporal} incorporates advanced architectures, Transformer~\cite{vaswani2017attention} and Multiresolution Graph Neural Network(MGNN)~\cite{hy2023multiresolution}, to capture spatio-temporal patterns in epidemic forecasting.

\subsection{LLM for Epidemic Forecasting}
Recently, LLMs have demonstrated their capacity to redefine various field as foundation models~\cite{liang2024foundation,gong2025agents}. 
The adaptation of LLMs for epidemic forecast is still in its early stages, and can be broadly categorized into two main lines: complex reasoning and time-series prediction.

\paragraph{Complex reasoning}
The research along this line focuses on leveraging the strong reasoning capabilities of LLMs.
PandemicLLM~\cite{du2024advancing} first reformulates the epidemic forecasting as a complex reasoning problem, incorporates textual policies and genomic surveillance data to enhance epidemic prediction. 
Sharing the same idea, Shah et al~\cite{shah2024infectious} integrate the climate data and textual epidemic data into an LLM-based epidemic prediction framework.
However, the complex factor data involved in epidemics are often scarce or confidential in real-world scenarios, limiting the utility of LLMs.

\paragraph{Time-series prediction}
Another line of studies attempt to predict epidemics via LLMs from the time-series perspective.
By modeling the epidemic transmission as time-series process, LLM-based forecasters of time-series~\cite{liu2024autotimes,zheng2025autocas} can be easily adapted.
Dey et al.~\cite{dey2024we} first introduce LLMs to epidemic prediction in the form of time-series forecasting.
They rigorously assess wether LLMs excels traditional statistical and machine learning models~\cite{kufel2020arima}, and confirms the feasibility of epidemic time-series foundation models based on LLMs~\cite{ansari2024chronos,feng2024only}.
Despite the relative ease of obtaining epidemic time-series data, such methods often overlook more complex spatio-temporal pattern factors influencing transmission~\cite{nguyen2023predicting}, such as the significant impact of population movement	human or mobility on the infection dissemination~\cite{panagopoulos2021transfer}.

\section{Preliminaries}
\subsection{Problem Definition}
In this paper, we formulate the epidemic forecasting tasks as a time-series prediction problem based on the human mobility network.
Let $X_t\in\mathbb{R}^{N\times F}$ denote the epidemiological features (historical infection cases in our paper) for $N$ regions at time $t$, where $F$ is the feature dimension.
Epidemic forecasting relies on the sequence of historical data $X_{1:T}=\{X_1,...,X_T\} \in \mathbb{R}^{T\times N \times F}$. 
It also incorporates mobility covariates $M_{1:T} \in \mathbb{R}^{T \times N \times N}$, where $(M_{t})_{ij}$ represents the human mobility from region $i$ to $j$.
We can interpret regions as nodes in a graph and the nonzero entries in 
$M_{1:T}$ as weighted edges to obtain a dynamic weighted graph $A_{1:T}\in \mathbb{R}^{T \times N \times N}$~\cite{panagopoulos2021transfer}. 
This graph then represents the potential pathways for epidemic transmission: infections can occur, in principle, only where there is nonzero population flow.
The objective is to forecast the case number for $N$ regions at the future time $T+h$ through a predictive model $f(\cdot)$:
\begin{equation}
    \hat{X}_{T+1:T+h} = f(X_{1:T}, A_{1:T}, M_{1:T}),
\end{equation}
where $h$ denotes the horizon time of epidemic forecasting. 

\subsection{Autoregressive Modeling}
Given a large collection of raw sequence data, we can employ a tokenizer to preprocess all of them into a 1D sequence.
This produces a dataset of tokens, $\{x_1,x_2,...,x_n\}$ where $n$ is the number of tokens.
We model the joint density $p(x)$ in autoregressive manner:
\begin{equation}
    p(x) = \prod_{i=1}^{n} p(x_i|x_{i-1}, x_{i-2}, ..., x_1, \theta),
\end{equation}
where $\theta$ denotes model parameters, which can be optimized by the target loss function.
Based on large-scale autoregressive pre-training, LLMs possess strong next-token-prediction capabilities~\cite{wang2022language}. 
Through tokenization and alignment with LLM architecture, multi-step generation and prediction can be implemented for corresponding data in specific domains~\cite{liu2023pre,liu2024autotimes,zheng2025autocas}. 

\section{Methodology}

\subsection{Framework Overview}
\begin{figure*}[htbp]
    \centering    \includegraphics[width=0.98\linewidth]{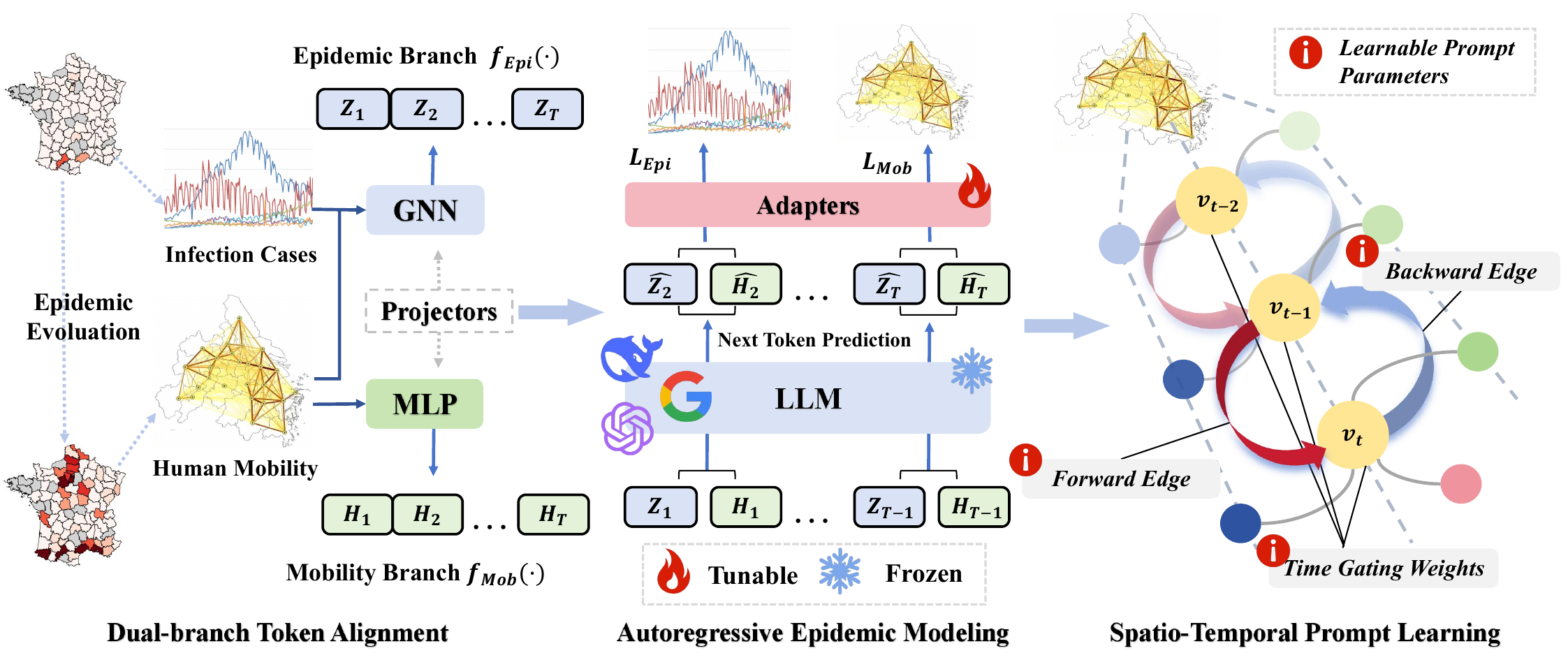}
    \caption{The overall framework of \name~consists of three modules: (1) dual-branch token alignment, (2) autoregressive epidemic modeling, and (3) spatio-temporal prompt learning.}
    \label{framework}
\end{figure*}

To apply LLMs into epidemic forecasting, we introduce \name~to unlock the potential of the LLM architecture in epidemic modeling.
The framework overview is illustrated in Figure~\ref{framework}, \name~consists of three main components: (1) dual-branch token alignment, (2) autoregressive epidemic modeling, (3) spatio-temporal prompt learning.
The dual-branch alignment module tokenizes the spatio-temporal epidemics to align with the LLM architecture. With the integration of autoregressive epidemic modeling, prompt learning techniques significantly enhance the LLM architecture to adapt to spatio-temporal epidemic forecasting.

\subsection{Dual-branch Token Alignment}
The key to adapting LLMs for epidemic forecasting lies in tokenizing epidemic dynamics~\cite{liu2024autotimes,zheng2025autocas,jin2023time}.
Different from nature language or naive time-series data, the complexity of real-world epidemics manifests in its spatio-temporal patterns~\cite{hy2022temporal,nguyen2023predicting}.
Considering the close interactions between epidemic progression and human mobility~\cite{panagopoulos2021transfer}, we design a dual-branch architecture to perform token-level alignment for epidemics.

\paragraph{Epidemic branch}
The epidemic branch focuses on the evolution dynamics of the epidemic.
Specifically, we temporally discretize the epidemic transmission process into $T$ time patches, and introduce a GNN to map the epidemiological features such as infection cases into tokens:
\begin{equation}
    Z_{1:T} = \text{GNN}(X_{1:T},A_{1:T}) \in \mathbb{R}^{T\times N \times D},
\end{equation}
where $D$ is consistent with the dimension of adopted LLMs. 
Intuitively, the introduced GNN can be seen as a $\text{Projector}(\cdot):\mathbb{R}^F \rightarrow \mathbb{R}^D$ to capture the spatio-temporal patterns for each timestep like nature language tokenizer~\cite{liu2024autotimes}.

\paragraph{Mobility branch}
While the epidemic branch accounts for human mobility, we still set up a dedicated mobility branch for fine-grained supervision due to the LLM generalization power.
Dual-branch collaboration facilitates multi-step joint epidemic prediction and paves the way for future epidemic foundation models with unified tasks.
The tokenization of mobility branch is similar to the epidemic branch, but we introduce a MLP as the projector based on the characteristics mobility data: 
\begin{equation}
    H_{1:T} = \text{MLP}(M_{1:T}) \in \mathbb{R}^{T \times N \times D}.
\end{equation}

After dual-branch token alignment, both epidemiological features and mobility dynamics are unified into the representation space of LLMs for collaborative modeling and further prediction. 

\subsection{Autoregressive Epidemic Modeling}
Based on large-scale autoregressive pre-training, prevalent LLMs can effectively predict the next token based on the preceding tokens~\cite{wang2022language}.
To fully unleash the potential of LLM architecture, we attempt to redefine the epidemic modeling via autoregressive paradigm.

\paragraph{Training phase}
For the epidemic branch, we feed the obtained tokens into the intermediate layers of LLM and perform next-token prediction for each patch:
\begin{equation}
    \{\hat{Z}_2,...,\hat{Z}_{T}\} = \text{LLM\_Layers}(\{Z_1,...,Z_{T-1} \}).
\end{equation}
After that, each predicted patch is mapped back by an $\text{Adapter}(\cdot):\mathbb{R}^D \rightarrow \mathbb{R}^F$ into the original input space for fine-grained supervision:
\begin{equation}
    \hat{X}_i = \text{Adapter}(\hat{Z}_i), i=2,...,T.
\end{equation}

Finally, each predicted patch is supervised by the token-wise ground truth to optimize the parameters of projector and adapter:
\begin{equation}
    \mathcal{L}_{Epi} = \frac{1}{N \times T} \sum_{n=1}^N \Vert X_i - \hat{X}_i \Vert_2, i=2,...,T.
\end{equation}
Sharing the same way, the mobility branch can also adopt the next token prediction for autoregressive modeling. 
Assuming the loss function of the mobility branch is $\mathcal{L}_{\text{Mob}} = \frac{1}{N\times T} \sum_{n=1}^N \Vert M_i - \hat{M}_i \Vert_2, i=2,...,T$, the final loss $\mathcal{L}_{\text{final}}$ of our framework during the training phase is:
\begin{equation}
    \mathcal{L}_{final} = \mathcal{L}_{Epi} + \lambda \mathcal{L}_{Mob}.
\end{equation}
where $\lambda$ is a weighting coefficient to balance the dual-branch losses.
It is worth noting that dual branches employ decoupled adapters to prevent task objective conflicts.
Moreover, we freeze the parameters of LLM, only tune the parameters of the light-weight projectors and adapters, significantly
reducing computational overhead and enabling quick task adaptation~\cite{zheng2025autocas}.


\paragraph{Inference phase}
At the inference phase, we seamlessly integrate the dual branches for epidemic forecasting. First, we obtain the spatial structure based on the mobility branch:
\begin{equation}
    \hat{A}_{T+i} = f_{Mob}(\hat{M}_{T+i-1}),
    i=1,...,h,
\end{equation}
where $f_{Mob}(\cdot)$ denotes the mobility branch that integrates the projector, LLM layers, and adapters.
After that, we further utilize the predicted spatial structure and epidemiological features to jointly forecast the future trend of epidemics:
\begin{equation}
    \hat{X}_{T+i} = f_{Epi}(\hat{X}_{T+i-1},\hat{A}_{T+i-1}),
    i=1,...,h,
\end{equation}
where $f_{Epi}(\cdot)$ denotes the epidemic branch. 
Notably, \name~integrates both human mobility and infection feature prediction, enabling predictions of arbitrary lengths by iterative multi-step generation.
Benefiting from the large-scale autoregressive pre-training and powerful next-token prediction capability, EpiLLM inherently excels at multi-step epidemic forecasting.

\subsection{Spatio-temporal Prompt Learning}
To further unlock the potential of LLMs in epidemic modeling, we introduce the prompt learning techniques from spatio-temporal perspective.
The core idea of prompt learning is to set learnable parameters to modify the input for specific task tuning~\cite{liu2023pre}.
Given a node $v_t$ representing a region at time $t$, with its mobility matrix denoted as $A_t$, we additionally incorporate the spatio-temporal dependency via a pair direction-aware edges $e^t_{forward}$ and $e^t_{backward}$ during the prompt phase:
 \begin{equation}
     e^t_{forward} = (v_{t-1},v_{t}), \ \ e^t_{backward} = (v_{t},v_{t-1}),
 \end{equation}
where the prompted edges follow the temporal directionality prior~\cite{yu2017spatio} and their weights are learnable.
Connecting regions to their previous time steps helps capture the evolution of dynamic systems and enhances the model perception of spatio-temporal patterns.
Each timestep within the token
window share a pair prompted edges to mitigate overfitting. 
Given the adjacency matrix $P_t$ with prompted edges $\{ e^t_{forward}, e^t_{backward}\}$ at time $t$, the prompted input $A_t^p$ can be expressed as:
$
    A_t^p = \bigcup_{j=t-w}^t A_j \oplus \bigcup_{j=t-w}^t P_j,
$
where $A^p_t$ denotes the prompted spatio-temporal structure at time $t$, $\oplus$ denotes matrix concatenation, and $w$ represents the token window length.
Specifically, we adopt the $\text{GNN}(\cdot)$ to tokenize the epidemics within a fixed-size window (token window) in the epidemic branch, and further combine it with a time gating mechanism to obtain the final tokens:
\begin{equation}
    Z_t = \sum_{k=t-w}^{t} \text{sigmoid}(\gamma_k) \ \odot \ \text{GNN}(X_k,A^p_k),
\end{equation}
where $\odot$ denotes the hadamard product and $\gamma_t$ represents the learnable gating weight at timestep $t$.
Intuitively, this prompt learning technique enhances the model's capability to capture spatio-temporal patterns of epidemic progression from a data-driven perspective.
More explanations regarding the spatio-temporal prompt learning and the prompt initialization strategy can be found in Appendix~\ref{app_prompt}.

\section{Experiments}
We perform thorough experiments on real-word COVID-19 datasets to evaluate \name~ and try to answer the following questions:
\noindent \textbf{RQ1:} How effective is \name~for epidemic forecasting?
\noindent \textbf{RQ2:} How key components of EpiLLM affect its performance?
\noindent \textbf{RQ3:} Does \name~exhibit the scaling behavior from LLMs?
\noindent \textbf{RQ4:} How does \name~achieves forecasting explainability and efficiency?

\subsection{Experimental Settings}
\paragraph{Datasets}
We evaluate the \name~using four real-world COVID-19 datasets~\cite{panagopoulos2021transfer}: England, France, Italy, and Spain.
The target is to predict the number of newly cases in specific regions of given countries.
Basic statistics and further dataset details are provided in Appendix~\ref{app_datasets}.

\paragraph{Setup and evaluation}
Following the previous studies~\cite{panagopoulos2021transfer,nguyen2023predicting,hy2022temporal}, we evaluate the proposed method in short-, mid- and long-term epidemic forecasting.
The autoregressive window corresponds to the common disease incubation, with settings of 3 days and 7 days. 
Correspondingly, the horizon window of prediction is set to \{3, 6, 7, 14\} days, where \{3, 7\} days adopt direct forecasting and \{6, 14\} days adopt multi-step forecasting. 
More details of data split in our implementation can be seen in Appendix~\ref{app_datasets}.
For evaluation metrics, we use Mean Absolute Error (MAE) and Root Mean Square Error (RMSE)  evaluate the performance (see in Appendix~\ref{metrics}).



\paragraph{Baselines}
To evaluate \name, we compare it with the state-of-the-art methods from three categories.
More baseline details can be found in Appendix~\ref{app_baseline}.

Statistical methods: AVG~\cite{hy2022temporal}, AVG\_WINDOW~\cite{hy2022temporal}, LAST\_DAY~\cite{hy2022temporal},
PROPHET~\cite{mahmud2020bangladesh}, ARIMA~\cite{kufel2020arima}.

Machine learning methods: LIN\_REG~\cite{kaur2022forecasting}, GP\_REG~\cite{ketu2021enhanced}, RAND\_REG~\cite{galasso2022random}, XGBOOST~\cite{fang2022application}.

Deep learning methods: LSTM~\cite{chimmula2020time},
MPNN~\cite{panagopoulos2021transfer}, MGNN~\cite{hy2022temporal}, MPNN+LSTM~\cite{panagopoulos2021transfer}, ATMGNN~\cite{nguyen2023predicting}.

Note that mechanistic models such as SIR and its variants are omitted since the datasets only include the number of case and does not involve infection rates, intervention policies, or other factors.

\paragraph{Implementation} 
We implement the \name~with the PyTorch framework on NVIDIA RTX 4090 GPU with 24GB of memory.
We adopt prevalent LLMs as the backbone of \name, incorporating models with varying parameter scales including GPT2~\cite{radford2019language}, DeepSeekR1~\cite{guo2025deepseek}, and GEMMA3~\cite{team2025gemma}, which can be downloaded from Huggingface\footnote{https://huggingface.co/}. 
In our implementation, we set the token window lengths to \{3, 7\} in the spatio-temporal prompt learning phase. 
Each region uses the historical number of new cases within a fixed window size as epidemiological features, where the feature window size is consistent with the token window length. 
We run the experiments 10 times to report the average results, and we employ the Adam optimizer~\cite{kingma2014adam} and adopt the early stopping strategy.

\begin{table*}
    \caption{The performance of direct epidemic forecasting for COVID-19. The best results of existing baselines are highlighted with \textbf{\textcolor{blue}{blue}}; the best results for EpiLLM are marked with \textbf{\textcolor{red}{red}}. The improvement(\%) is calculated based on the aforementioned two results.
    Experimental results have passed the statistical significance tests.
    }
    \label{results}
    \resizebox{\textwidth}{!}{
    \begin{tabular}{c | c c | c c | c c | c c | c c | c c| c c | c c }
    \toprule
    \multirow{4}{*}{\textbf{Model}} & 
    \multicolumn{4}{c|}{\textbf{England}} &
    \multicolumn{4}{c|}{\textbf{France}} &
    \multicolumn{4}{c|}{\textbf{Italy}} &
    \multicolumn{4}{c}{\textbf{Spain}} \\ 
    \cmidrule(r){2-17}
    & \multicolumn{2}{c|}{3 days} & \multicolumn{2}{c|}{7 days} & \multicolumn{2}{c|}{3 days} & \multicolumn{2}{c|}{7 days} & \multicolumn{2}{c|}{3 days} & \multicolumn{2}{c|}{7 days}  & \multicolumn{2}{c|}{3 days} & \multicolumn{2}{c}{7 days} \\
    \cmidrule(r){2-17}
    &RMSE & MAE & RMSE & MAE & RMSE & MAE & RMSE & MAE & RMSE & MAE & RMSE & MAE & RMSE & MAE & RMSE & MAE \\
    \midrule
        AVG & 11.39 & 8.15 & 11.77 & 8.50 & 14.41 & 7.65 & 14.22 & 7.55 & 42.80 & 21.13 & 42.8 & 21.13 & 111.19 & 48.71 & 122.82 & 52.69  \\ 
        AVG\_WINDOW & 8.79 & 6.33 & 10.87 & 7.94 & 9.47 & 5.24 & 10.14 & 5.69 & 33.48 & 17.69 & 33.48 & 17.69 & 59.57 & 32.56 & 79.83 & 40.09  \\ 
        LAST\_DAY & 10.45 & 7.12 & 10.49 & 7.33 & 13.94 & 7.29 & 9.84 & 5.05 & 41.99 & 21.21 & 41.99 & 21.21 & 63.98 & 35.20 & 70.57 & 37.63  \\ 
        PROPHET & 15.67  & 6.86  & 24.45  & 11.50  & 11.44  & 6.25  & 20.66  & 9.74  & 32.23  & 18.88  & 35.82  & 18.96  & 127.49  & 41.36  & 80.37   & 75.86  \\ 
        ARIMA & 15.30  & 6.59  & 10.12  & 8.52  & 7.41  & 4.66  & 7.59  & 4.67  & 52.00  & 24.77  & 49.28  & 20.15  & 40.79  & 20.21  & 66.51   & 40.54  \\
        \midrule
        LIN\_REG & 13.40  & 9.67  & 16.87  & 15.40  & 5.34  & 2.99  & 11.57  & 8.21  & 42.49  & 23.07  & 46.00  & 21.95  & 58.67  & 34.47  & 85.72   & 62.34  \\ 
        GP\_REG & 14.05  & 11.01  & 17.25  & 12.66  & 3.55  & 2.36  & 6.31  & 4.11  & 58.22  & 26.92  & 55.56  & 29.17  & 56.43  & 31.00  & 92.34   & 51.28  \\ 
        RAND\_FOREST & 7.44  & 5.23  & 9.51  & 6.82  & 5.99  & 2.78  & 5.13  & 4.08  & 27.71  & 14.98  & 34.42  & 17.09  & 53.88  & 33.38  & 57.72   & 37.05  \\ 
        XGBOOST & 8.24  & 5.66  & 10.12  & 7.94  & 6.73  & 2.36  & 5.64  & 4.29  & 36.69  & 17.86  & 35.02  & 16.99  & 38.36  & 24.17  & 67.14   & 38.18  \\ 
        \midrule
        LSTM & 7.77  & 5.48  & 9.39  & 7.17  & 5.56  & 3.20  & 6.04  & 3.96  & 28.53  & 13.31  & 31.04  & 18.46  & 37.73  & 20.93  & 57.80   & 44.25  \\ 
        MPNN & 7.10  & 4.89  & 7.81  & 6.76  & 4.68  & 3.15  & 5.88  & 3.81  & 24.91  & 13.12  & 27.57  & 14.78  & 36.88  & 21.72  & 64.15   & 39.91  \\ 
        MGNN & 7.15  & 5.06  & 8.04  & 6.51  & 3.62  & 2.83  & 5.47  & 4.57  & \textbf{\textcolor{blue}{24.53}}  & 13.77  & 27.64  & 14.92  & 37.25  & 20.22  & 65.53   & 42.35  \\ 
        MPNN+LSTM & 7.20  & 4.95  & \textbf{\textcolor{blue}{7.45}}  & \textbf{\textcolor{blue}{5.64}}  & 3.58  & 2.78  & 5.06  & 4.64  & 29.95  & 13.06  & \textbf{\textcolor{blue}{27.28}}  & \textbf{\textcolor{blue}{14.73}}  & 34.16  & \textbf{\textcolor{blue}{19.95}}  & 57.38   & 35.03  \\ 
        ATMGNN & \textbf{\textcolor{blue}{5.77}} & \textbf{\textcolor{blue}{3.97}} & 7.55 & 5.77 & \textbf{\textcolor{blue}{3.45}} & \textbf{\textcolor{blue}{2.25}} & \textbf{\textcolor{blue}{4.65}} & \textbf{\textcolor{blue}{3.79}} & 25.09 & \textbf{\textcolor{blue}{12.99}} & 27.47 & 15.88 & \textbf{\textcolor{blue}{32.12}} & 21.85 & \textbf{\textcolor{blue}{55.90}} & \textbf{\textcolor{blue}{30.79}}  \\ 
    \midrule
    EpiLLM-GPT2 & 5.41 & 3.83 & 6.22 & 4.39 & 3.41 & 2.11 & 4.16 & 3.75 & 22.64 & 12.97 & 26.06 & 14.35 & 26.85 & 19.94 & 40.46 & 28.31  \\
    EpiLLM-DeepSeekR1 & 5.37 & 3.71 & 6.19 & 4.28 & 3.22 & 2.07 & 4.18 & 3.77 & \textbf{\textcolor{red}{21.65}} & 12.40 & 24.04 & 14.18 & 26.72 & 19.12 & 39.78 & 27.16  \\ 
    EpiLLM-GEMMA3 & \textbf{\textcolor{red}{5.30}} & \textbf{\textcolor{red}{3.63}} & \textbf{\textcolor{red}{6.02}} & \textbf{\textcolor{red}{4.25}} & \textbf{\textcolor{red}{3.07}} & \textbf{\textcolor{red}{1.91}} & \textbf{\textcolor{red}{3.25}} & \textbf{\textcolor{red}{3.27}} & 21.67 & \textbf{\textcolor{red}{11.87}} & \textbf{\textcolor{red}{22.26}} & \textbf{\textcolor{red}{14.09}} & \textbf{\textcolor{red}{26.08}} & \textbf{\textcolor{red}{18.67}} & \textbf{\textcolor{red}{38.92}} & \textbf{\textcolor{red}{26.32}}  \\
    \midrule
    Improvement(\%) & \textbf{7.62} & \textbf{8.56} & \textbf{19.19} & \textbf{24.64} & \textbf{9.91} & \textbf{15.11} & \textbf{17.30} & \textbf{13.72} & \textbf{11.74} & \textbf{9.11} & \textbf{17.67} & \textbf{4.34} & \textbf{18.80} & \textbf{6.41} & \textbf{30.38} & \textbf{14.51} \\
    \bottomrule
    \end{tabular}  
    }
    \end{table*}
\begin{table*}
    \caption{The performance of multi-step epidemic forecasting for COVID-19. The best results of EpiLLM are highlighted with \textbf{\textcolor{red}{red}}. Experimental results have passed the statistical significance tests.
    }
    \label{autoregressive}
    \resizebox{\textwidth}{!}{
    \begin{tabular}{c | c c | c c | c c | c c | c c | c c| c c | c c }
    \toprule
    \multirow{4}{*}{\textbf{Model}} & 
    \multicolumn{4}{c|}{\textbf{England}} &
    \multicolumn{4}{c|}{\textbf{France}} &
    \multicolumn{4}{c|}{\textbf{Italy}} &
    \multicolumn{4}{c}{\textbf{Spain}} \\ 
    \cmidrule(r){2-17}
    & \multicolumn{2}{c|}{6 days} & \multicolumn{2}{c|}{14 days} & \multicolumn{2}{c|}{6 days} & \multicolumn{2}{c|}{14 days} & \multicolumn{2}{c|}{6 days} & \multicolumn{2}{c|}{14 days}  & \multicolumn{2}{c|}{6 days} & \multicolumn{2}{c}{14 days} \\
    \cmidrule(r){2-17}
    &RMSE & MAE & RMSE & MAE & RMSE & MAE & RMSE & MAE & RMSE & MAE & RMSE & MAE & RMSE & MAE & RMSE & MAE \\
    \midrule
    EpiLLM-GPT2 & 6.92 & 5.20 & \textbf{\textcolor{red}{7.75}} & \textbf{\textcolor{red}{6.02}} & 3.62 & 2.37 & 5.13 & 4.03 & 30.78 & 14.62 & 43.05 & 26.74 & 35.40 & 23.85 & 56.85 & 37.88  \\ 
    EpiLLM-DeepSeekR1 & \textbf{\textcolor{red}{6.04}} & \textbf{\textcolor{red}{4.30}} & 8.34 & 6.40 & 3.84 & \textbf{\textcolor{red}{2.07}} & 5.26 & 4.24 & 25.49 & 18.00 & 42.78 & 24.10 & \textbf{\textcolor{red}{29.92}} & \textbf{\textcolor{red}{20.39}} & 55.43 & \textbf{\textcolor{red}{36.02}}  \\ 
    EpiLLM-GEMMA3 & 6.56 & 4.52 & 8.77 & 6.60 & \textbf{\textcolor{red}{3.53}} & 2.18 & \textbf{\textcolor{red}{5.05}} & \textbf{\textcolor{red}{3.98}} & \textbf{\textcolor{red}{24.33}} & \textbf{\textcolor{red}{14.57}} & \textbf{\textcolor{red}{36.11}} & \textbf{\textcolor{red}{21.50}} & 32.15 & 21.40 & \textbf{\textcolor{red}{49.14}} & 36.92  \\ 
    \bottomrule
    \end{tabular}  
    }
    \end{table*}
    
\subsection{Performance Comparison (RQ1) }
\label{rq1}

We evaluate the effectiveness of \name~from two perspectives: direct forecasting and multi-step forecasting, and provide detailed case studies in Appendix~\ref{app_casestudy}.

\paragraph{Direct forecasting} 
Direct forecasting refers to performing single-step prediction for the next horizon window, which most effectively demonstrates a model's capability in epidemic modeling.
We compare \name~with 14 representative baselines across four real-world COVID-19 datasets, and experimental results are shown in Table~\ref{results}.
Overall, we can observe that \name~demonstrates significantly superior performance compared to other baselines in direct epidemic forecasting.
Particularly on the Spain dataset, \name~ achieves an impressive 30.38 \% improvement over the best-performing baseline on RMSE, highlighting its effectiveness.
Enhanced forecasting performance stems from the LLM's advanced next-token prediction and holistic spatio-temporal modeling, while EpiLLM retains seamless integration with existing LLM architectures.
Moreover, deep learning methods show competitive performance over statistical methods and machine learning methods, where ATMGNN excels due to the integration of spatio-temporal epidemic patterns.

\paragraph{Multi-step forecasting}
To evaluate the multi-step forecasting capability of \name, we set the horizon window to 6/14 days and perform autoregressive prediction.
As observed in Table~\ref{autoregressive}, EpiLLM-GEMMA demonstrates superior multi-step generation capabilities compared to EpiLLM-GPT2 and EpiLLM-DeepSeekR1.
This observation aligns with the conclusions in technical report of GEMMA3~\cite{team2025gemma}, where this advanced architecture demonstrates the capability to reduce error accumulation in long-sequence generation.
Thanks to the generation of mobility covariates, \name~can jointly combine infection cases and human mobility to predict in a continuous manner. 
Other baselines fail to perform multi-step forecasting due to the absence of future mobility prediction, highlighting the importance of the dual-branch alignment module and the strong generalization of LLM architecture.

\subsection{Ablation Study (RQ2)}

To validate the effectiveness of each component in \name, we  conduct ablation studies via the full models with 8 variants:
(1) \textit{Graph2MLP} uses only epidemiological features without human mobility, following the pipeline in \textbf{AutoTimes}\cite{liu2024autotimes}.
(2) \textit{Time2Aver} removes the time gating mechanism in prompt learning and replaces it with the average operation.
(3) \textit{Time2Last} removes the time gating mechanism and adopt embeddings at the last timestamp.
(4) \textit{w/o LLM} removes the LLM and directly feed tokens to the adapter.
(5) \textit{LLM2MLP} replaces the LLM with a MLP block.
(6) \textit{LLM2RNN} replaces the LLM with a RNN block.
(7) \textit{LLM2Trans} replaces the LLM with a vanilla Transformer block.
As shown in Figure~\ref{ablation}, \name~outperforms other variants without integrated spatial or temporal patterns while effectively leveraging the powerful generalization capabilities of LLMs. 
Additional ablation study details are provided in the Appendix~\ref{app_ablation}, including EpiLLM's effectiveness in predicting human mobility and its autoregressive modeling capabilities.

\begin{figure}[htbp]
\label{ablation}
\centering
\includegraphics[width=0.88\linewidth]{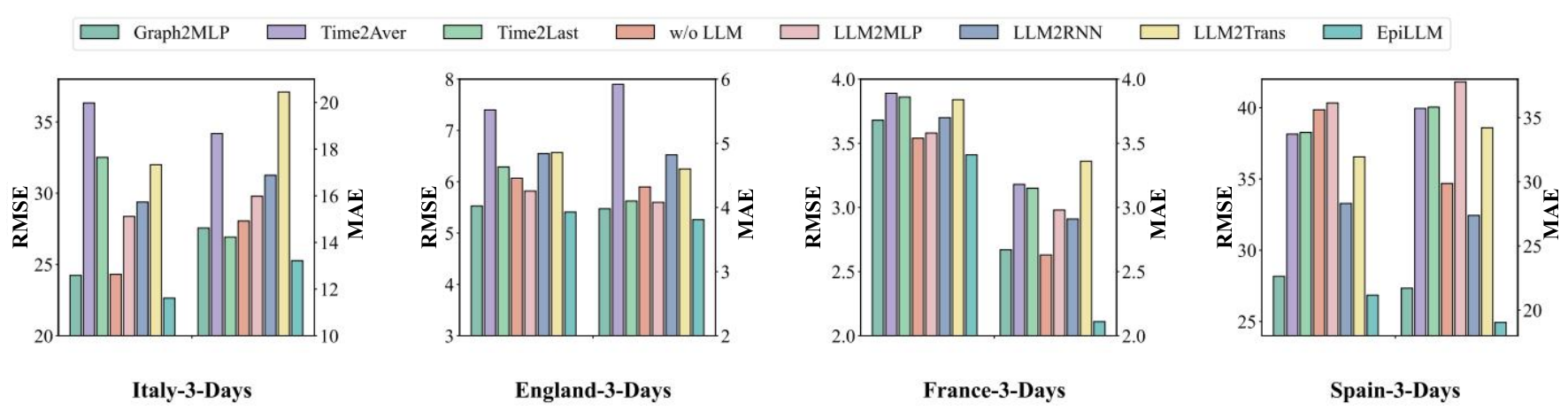}
\caption{Ablation study of EpiLLM for epidemic forecasting.}
\label{ablation}
\end{figure}
    
\subsection{Scaling Behavior (RQ3)}
\begin{figure}[htbp]
    \centering
    \begin{subfigure}[b]{0.26\linewidth}
        \centering
        \includegraphics[width=\linewidth]{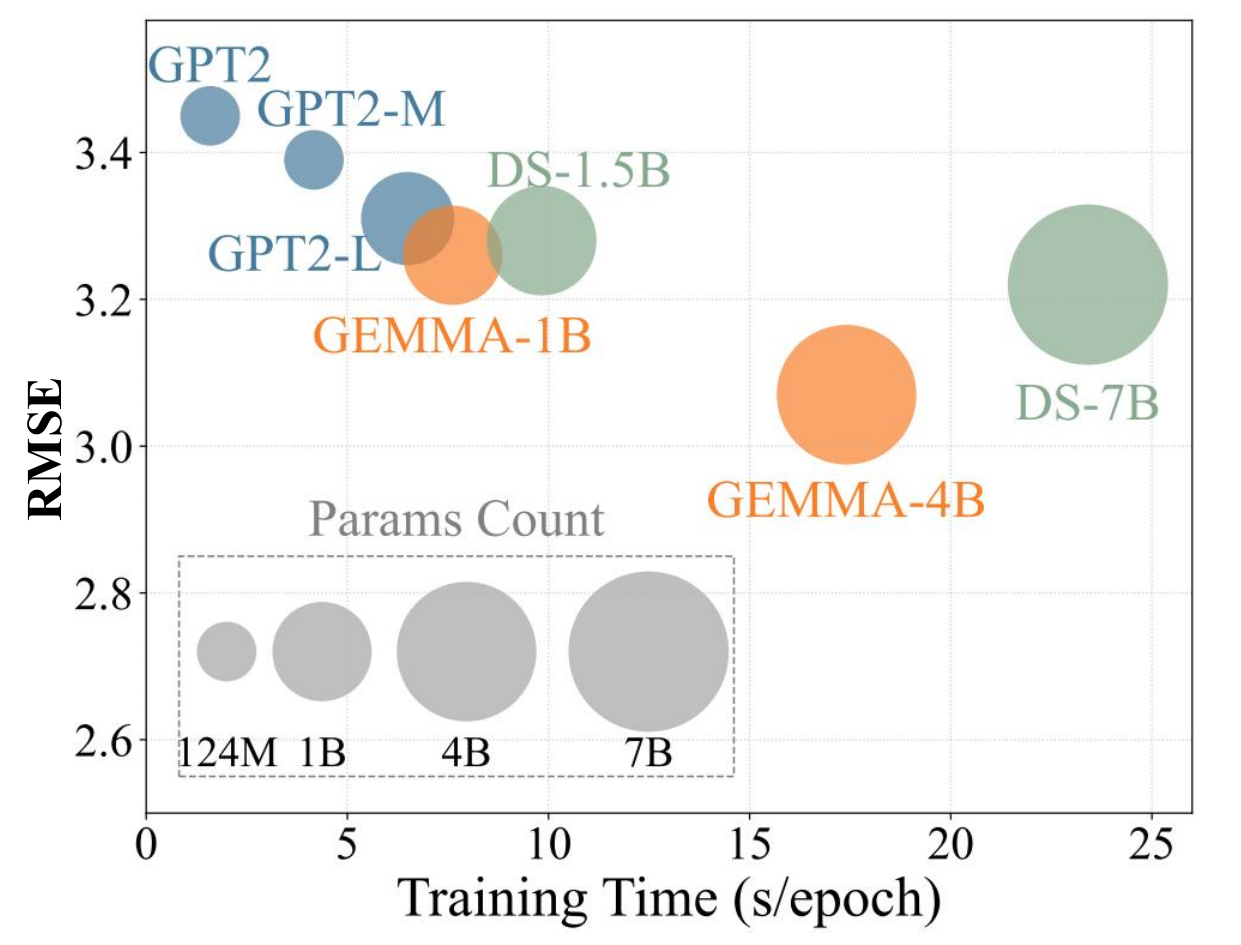}
        \caption{France}
    \end{subfigure}
    \hspace{1cm}
    \begin{subfigure}[b]{0.26\linewidth}
        \centering
        \includegraphics[width=\linewidth]{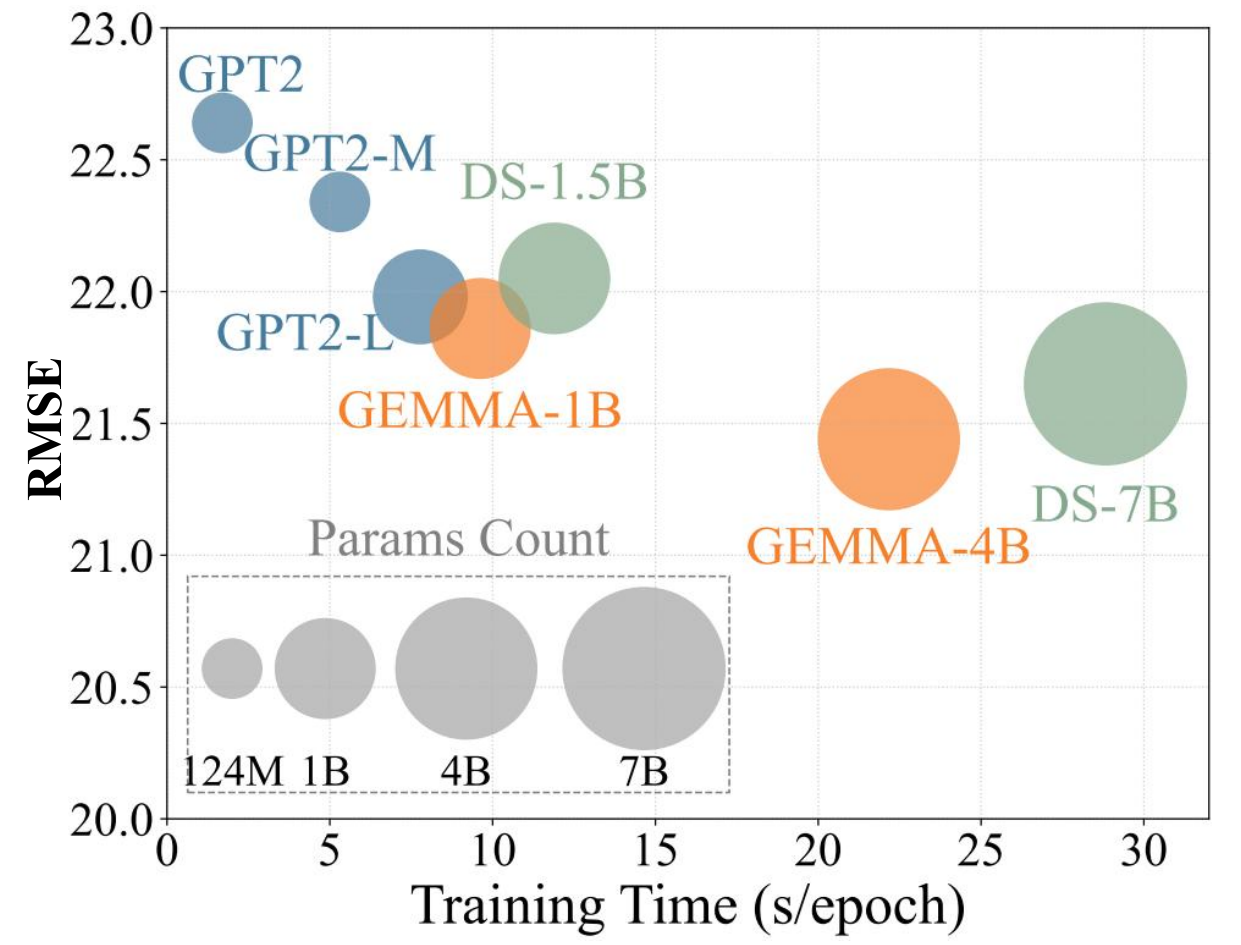}
        \caption{Italy}
    \end{subfigure}
    
    \caption{Scaling behavior of EpiLLM on France and Italy datasets.}
    \label{scaling}
\end{figure}

Scaling behavior refers to how foundation models systematically improve their performance as computational factors are increased.
This phenomenon is characterized by predictable relationships between model capabilities and scaling variables.
Here, we explore the scaling trends of \name~in epidemic forecasting and evaluate each adapted LLM predictor from three perspectives: forecasting performance, training speed, and parameter count.
In Figure~\ref{scaling}, we observe that the scaling phenomenon is particularly prominent in the GPT2 and GEMMA model family: the forecasting performance (measured by RMSE) of models exhibits consistent improvement with increasing parameter scale, albeit at the expense of greater computational demands, as evidenced by prolonged training durations.
The specific number of LLM parameters in this experiment can be referred to in Table~\ref{efficiency}.
Notably, DeepSeekR1-7B (DS-7B) with larger scale shows slightly inferior predictive performance compared to GEMMA-4B, which may be attributed to the more advanced architecture of GEMMA.

\subsection{Explainability and Efficiency Analysis (RQ4)}

\begin{figure}[htbp]
    \centering
    \begin{subfigure}[b]{0.21\linewidth}
    \centering
    \includegraphics[width=\linewidth]{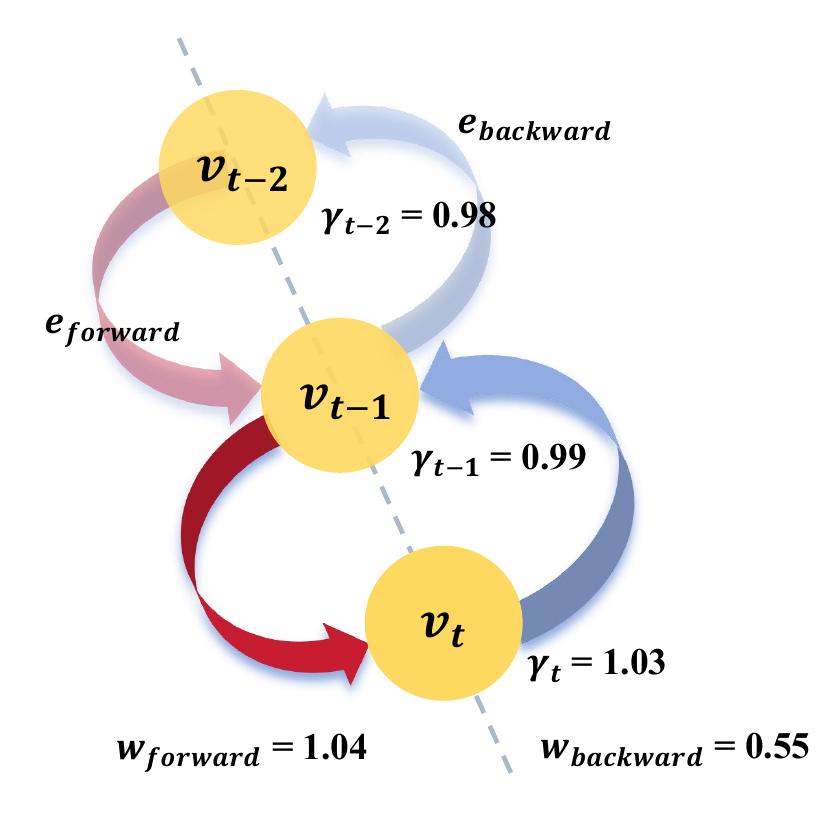}
    \caption{England-3-days}
    \end{subfigure}
    \begin{subfigure}[b]{0.21\linewidth}
    \centering
    \includegraphics[width=\linewidth]{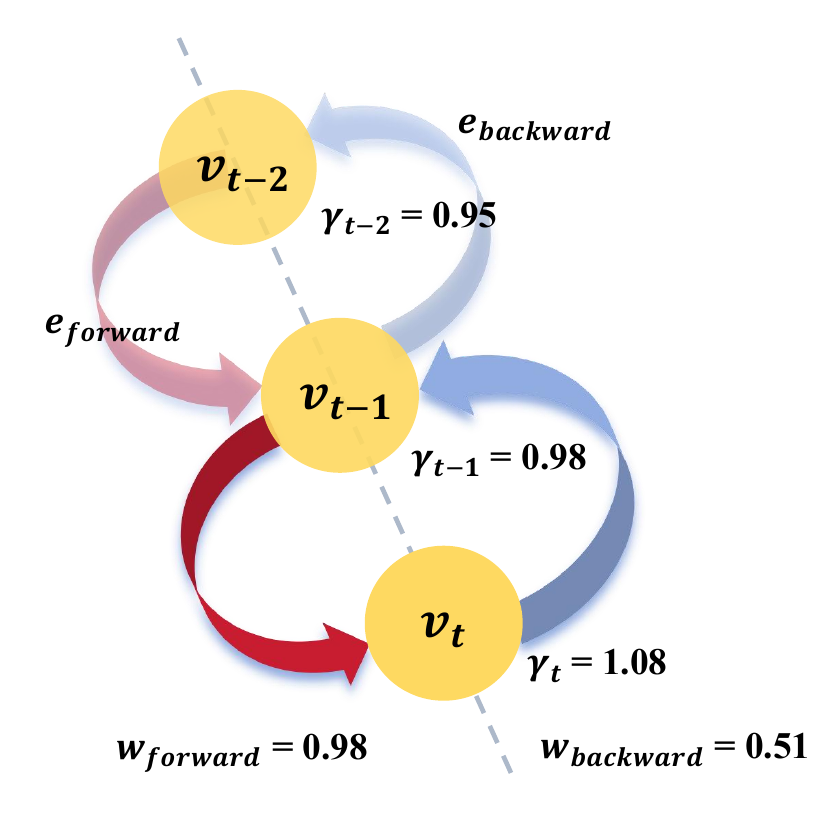}
     \caption{France-3-days}
    \end{subfigure}
    \begin{subfigure}[b]{0.21\linewidth}
    \centering
    \includegraphics[width=\linewidth]{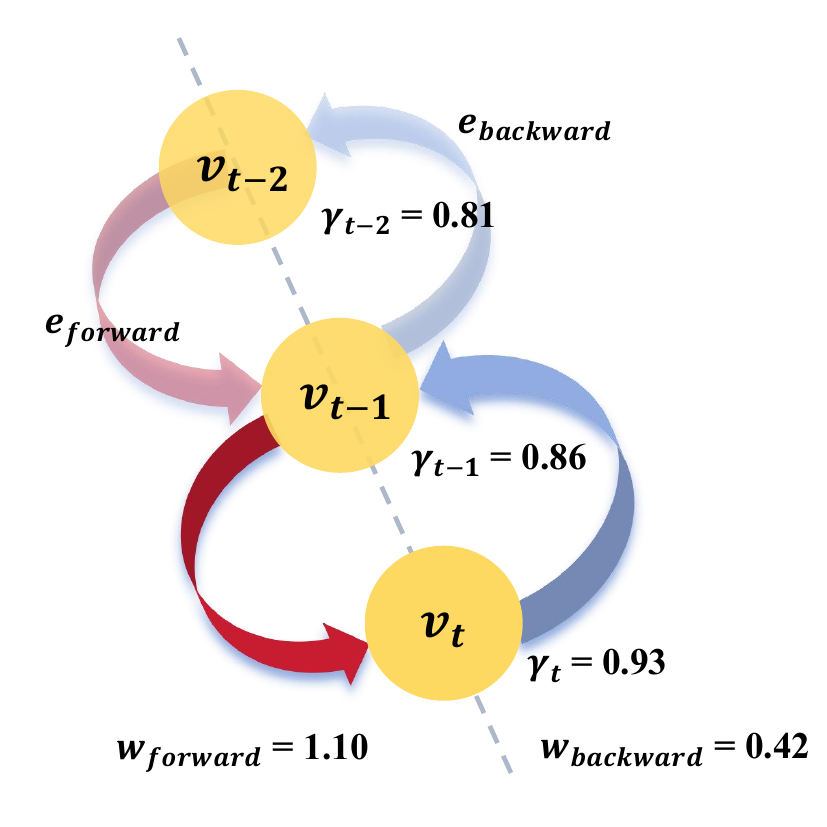}
     \caption{Italy-3-days}
    \end{subfigure}
    \begin{subfigure}[b]{0.21\linewidth}
    \centering
    \includegraphics[width=\linewidth]{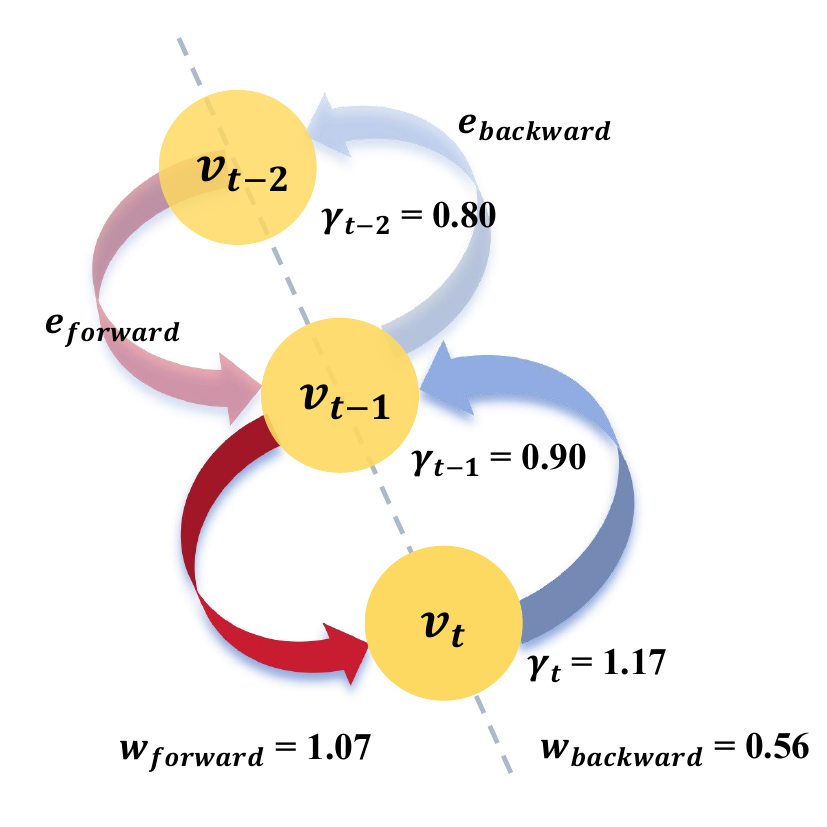}
     \caption{Spain-3-days}
    \end{subfigure}
    \caption{Prompt visualization of \name.}
    \label{prompt}
\end{figure}

\paragraph{Prompt visualization}
To evaluate the prompt explainability, we visualize the weights of direction-aware edges and time gating weights.
In Figure~\ref{prompt}, the time gating weights progressively increase over time, demonstrating \name~places greater emphasis on the current timestep and effectively captures the temporal dependencies.
For direction-aware edges, the trainable forward weight exceeds the backward one, adhering to the temporal directionality assumption.
During the prompting process, the gating weights and direction-aware edges synergistically enhance the spatio-temporal modeling while maintaining explainability in \name.
More experimental results can be found in Appendix~\ref{app_prompt}.

\paragraph{Parameter efficiency}
To evaluate the efficiency of \name, we focus on the computational efficiency and analyze its parameter utilization.
As observed in Table~\ref{efficiency}, the trainable parameters in EpiLLM constitute a minimal portion of the overall framework.
With the scaling up of backbone, total parameters of \name~increases significantly, leading to markedly improved performance, while the proportion of trainable parameters progressively decreases, highlighting its efficiency.

\begin{table}[htbp]
    \scriptsize
    \centering
    \caption{The statistics of parameter utilization in \name.}
    \label{efficiency}
    \begin{tabular}{c c c c c c c c c c}
    \toprule
    \textbf{Model} & GPT2 & GP2-M & GP2-L & DeepSeekR1-1.5B & DeepSeekR1-7B & GEMMA3-1B & GEMMA3-4B\\
    \midrule
    \textbf{Trainable Para.} & 727K & 930K & 1.13M & 1.33M & 2.93M & 1.03M & 2.14M\\
    \textbf{Total Para.} & 125M & 355M & 775M & 1.78B & 7.62B & 1.00B & 4.30B\\
    \textbf{Ratio} & 0.58\% & 0.26\% & 0.14\% & 0.07\% & 0.03\% & 0.10\% & 0.04\%\\
    \bottomrule
    \end{tabular}
\end{table}

\section{Conclusion}
In this paper, we present a novel framework that repurposes LLMs as real-world epidemic forecasters.
The introduced dual-branch alignment module tokenizes spatio-temporal epidemics to fit the LLM architecture. Integrated with autoregressive modeling, prompt learning further enhances the LLM adaptation to spatio-temporal epidemic forecasting.
Extensive experiments demonstrate the superior performance of \name, and it exhibits the scaling behavior empowered by LLMs.
In future research, we are attempting to develop a multi-modal foundation model for epidemic forecasting, as well as addressing potential security  threats and ethical controversies of LLMs in public health applications.

\bibliographystyle{unsrt}
\bibliography{ref}

\newpage
\appendix

\section{Further Details of Datasets}
\label{app_datasets}

\begin{table}[htbp]
    \scriptsize
    \centering
    \caption{The statistics information of epidemic forecasting datasets for COVID-19 .}
    \label{datasets}
    \begin{tabular}{c c c c c c}
    \toprule
    \textbf{Dataset} & England & France & Italy & Spain \\
    \midrule
    \textbf{Period} & 2020-03-13 – 2020-05-12 & 2020-03-10 – 2020-05-12 & 2020-02-24 – 2020-05-12 & 2020-03-12 – 2020-05-12 \\
    \textbf{Regions} & 129 & 81 & 105 & 34 \\
    \textbf{Avg. Cases} & 16.7 & 7.5 & 25.65 & 61\\
    \bottomrule
    \end{tabular}
\end{table}

\paragraph{Datasets construction}
COVID-19 is a newly identified virus in Wuhan, China in December 2019, which is a disease caused by severe acute respiratory syndrome coronavirus 2 or SARS-Cov-2 and closely related to bat coronaviruses, pangolin coronaviruses, and SARS-CoV.
In this paper, we focus on the epidemics of COVID-19 in European countires: England, France, Italy, and Spain.
The number of reported cases in the regions of these four countries is collected from open-source github repository\footnote{https://dataforgood.fb.com/tools/disease-prevention-maps/}.  
The human mobility data is collected from mobile devices with the Facebook App installed and location history settings enabled, which can be download from the github repository\footnote{https://github.com/geopanag/pandemic\_tgnn}.
The raw time-series dataset comprises tri-daily recordings (specifically at midnight, morning, and afternoon intervals) that quantify population movement volumes between regions during each corresponding diurnal phase.
We reused preprocessed time-series data from prior studies, where the three daily values were further aggregated into a single metric representing mobility between two regions.
The observation period initiates from the first date with synchronized mobility-case data. Exclusions applied to regions with: (i) no detected cases, or (ii) unlinkable Facebook mobility records.
Basic statistics of datasets are summarized in Table~\ref{datasets}.

\paragraph{Datasets splits}
Considering the characteristics of epidemic forecasting tasks—rapid outbreak and fast transmission—their spatiotemporal sequences typically span approximately 60 days.
Conventional cross-validation methods are incompatible with autoregressive prediction requirements, we adopt a temporally ordered dataset partitioning strategy, which better aligns with real-world epidemic transmission scenarios.
Specifically, the last \{3, 6 (autoregressive), 7, 14 (autoregressive)\} days of the sequence are reserved as the test set, while the \{3, 7\} days immediately preceding the test set serve as the validation set, with the remaining data allocated to the training set.

\section{Evaluation Metrics}
\label{metrics}
Let $y$ represents the ground truth, and $\hat{y}$ represents the predicted result in the horizon time $h$. 
The evaluation metrics for one region we used in this paper are defined as follows:
\paragraph{Mean Absolute Error (MAE) }

\begin{equation}
    MAE(y,\hat{y}) = \frac{1}{h} \sum_{t = T+1}^{T+d} \vert y_t - \hat{y}_t \vert
\end{equation}

\paragraph{Root Mean Square Error (RMSE) }
\begin{equation}
    RMSE(y,\hat{y}) = \sqrt{\frac{1}{h} \sum_{t = T+1}^{T+d}  ( y_t - \hat{y}_t )^2}
\end{equation}

\section{Further Details of Baselines}
\label{app_baseline}

To evaluate \name, we conducted a comparative analysis with 14 leading-edge models
in the domain of epidemic forecasting. 
The models we benchmark against are as follows:

AVG~\cite{hy2022temporal}: The average number of reported cases for each region up to the time of the test day.

AVG\_WINDOW~\cite{hy2022temporal}: The average number of reported cases for each region in the past horizon days.

LAST\_DAY~\cite{hy2022temporal}: The number of reported cases for each region in the previous day is used for prediction.

PROPHET~\cite{mahmud2020bangladesh}: A time-series model where the input is the history of entire reported cases for each region, which is widely used in epidemic forecasting.

ARIMA~\cite{kufel2020arima}: An autoregressive moving average model for time-series forecasting, which the input is similar to PROPHET.

LIN\_REG~\cite{kaur2022forecasting}: Given the history of reported cases for each region as input, ordinary least squares linear regression is used to fit the line of cases on the training sets to forecast the future epidemic trend.

GP\_REG~\cite{ketu2021enhanced}: A non-parametric based regression model commonly used for time-series forecast that implements the Gaussian processes. 

RAND\_FOREST~\cite{galasso2022random}: A random forest regression model that produces epidemic forecasting using decision trees, with multiple trees built based on the training sets to best average the final results.

XGBOOST~\cite{fang2022application}: An enhanced version of andom forest regression model for epidemic forecasting via gradient boosting.

LSTM~\cite{chimmula2020time}: Given the sequence of reported cases for each region as inputs, a two-layer long short-term memory network is used for prediction.

MPNN~\cite{panagopoulos2021transfer}: Given the time-series data of reported cases as inputs, a message-passing neural network~\cite{gilmer2017neural} with separate layers for each day.

MGNN~\cite{hy2022temporal}: Similar to MPNN, a message-passing neural network is enhanced with multiple graph resolutions and adaptive clustering scale for different regions. 

MPNN+LSTM~\cite{panagopoulos2021transfer}: A hybrid deep learning model for epidemic forecasting, where MPNN extracts spatial dependencies among regions, while LSTM captures the temporal dynamics of the epidemic.

ATMGNN~\cite{nguyen2023predicting}: A hybrid deep learning model for epidemic forecasting, where multiple resolution GNN~\cite{hy2023multiresolution} are combined with Transformers~\cite{vaswani2017attention} for modeling the epidemics.

\section{Ablation Study}

Here, we supplement more details of the ablation experiments. 
Human mobility prediction constitutes the core component of our framework, as its performance directly determines whether models integrating external human mobility knowledge can achieve effective multi-step forecasting. 
Beyond the demonstrated superiority of EpiLLM in direct prediction reported in the main text, we systematically validate the effectiveness of integrated human mobility through ablation studies. 
Specifically, for multi-step forecasting, we design 3 model variants: 
(1) \textit{Graph2MLP} uses only epidemiological features without human mobility, following the pipeline in AutoTimes.\cite{liu2024autotimes}. 
(2) \textit{Adj2Aver} removes the human mobility prediction module, substituting it with averaged adjacency matrices from historical time steps within the window. 
(3) \textit{Adj2Last} eliminates the human mobility prediction module and directly reuses the adjacency matrix from the preceding prediction step. 
As shown in Figure~\ref{ablation} and Table ~\ref{ablation2}, experimental results demonstrate EpiLLM's exceptional direct and multi-step forecasting capability, with ablation studies yielding key findings:

\textit{Graph2MLP} exhibits mediocre performance across datasets due to its disregard for spatial effects in human mobility. 
\textit{Adj2Aver} fails to consider temporal directionality priors~\cite{yu2017spatio}, neglecting important spatio-temporal patterns of the epidemic through naive averaging aggregation, thus achieving the poorest performance. \textit{Adj2Last} captures only immediate temporal dependencies while neglecting long-term spatio-temporal patterns, resulting in subpar outcomes. 
The above ablation experiments demonstrate the importance of integrating human mobility into the EpiLLM framework, while also highlighting that dual-branch collaborative prediction of disease dynamics and human mobility is a key condition for achieving multi-step epidemic forecasting.
Replacing the LLM backbone with trainable MLP block leads to significant performance degradation, demonstrating the importance of the autoregressive modeling paradigm for EpiLLM.
Replacing the LLM backbone with trainable RNN, Transformer block leads to suboptimal performance, indicating that the LLM architecture, after large-scale autoregressive pre-training, possesses strong autoregressive generation capabilities that are well-suited for spatio-temporal epidemic prediction tasks.
It is worth noting that the LLM-free variant 
\textit{(w/o LLM}) demonstrates acceptable performance when processing tokens directly through adapters, which can be attributed to the inherent predictive potential of our spatio-temporal prompt learning design.
\label{app_ablation}
\begin{table*}
    \caption{Multi-step forecasting ablation study of EpiLLM.}
    \label{ablation2}
    \resizebox{\textwidth}{!}{
    \begin{tabular}{c | c c | c c | c c | c c | c c | c c| c c | c c }
    \toprule
    \multirow{4}{*}{\textbf{Type}} & 
    \multicolumn{4}{c|}{\textbf{England}} &
    \multicolumn{4}{c|}{\textbf{France}} &
    \multicolumn{4}{c|}{\textbf{Italy}} &
    \multicolumn{4}{c}{\textbf{Spain}} \\ 
    \cmidrule(r){2-17}
    & \multicolumn{2}{c|}{6 days} & \multicolumn{2}{c|}{14 days} & \multicolumn{2}{c|}{6 days} & \multicolumn{2}{c|}{14 days} & \multicolumn{2}{c|}{6 days} & \multicolumn{2}{c|}{14 days}  & \multicolumn{2}{c|}{6 days} & \multicolumn{2}{c}{14 days} \\
    \cmidrule(r){2-17}
    &RMSE & MAE & RMSE & MAE & RMSE & MAE & RMSE & MAE & RMSE & MAE & RMSE & MAE & RMSE & MAE & RMSE & MAE \\
    \midrule
    Graph2MLP & 7.03 & 5.63 & 8.22 & 5.60 & 4.05 & 3.37 & 6.48 & 3.56 & 39.71 & 24.25 & 44.887 & 28.51 & 47.18 & 25.13 & 76.85 & 47.88  \\ 
    Adj2Aver & 9.64 & 7.06 & 10.46 & 7.85 & 4.02 & 2.95 & 6.08 & 4.96 & 39.46 & 18.62 & 51.68 & 30.34 & 42.64 & 27.27 & 60.19 & 38.95 \\ 
    Adj2Last & 7.68 & 5.94 & 8.89 & 6.32 & 3.92 & 2.86 & 5.79 & 4.45 & 36.97 & 16.81 & 48.13 & 28.64 & 39.79 & 24.67 & 59.76 & 28.27 \\ 
    \textbf{EpiLLM} & \textbf{6.92} &\textbf{ 5.2} & \textbf{7.75} & \textbf{6.02} & \textbf{3.62} & \textbf{2.37 }& \textbf{5.13} & \textbf{4.03} & \textbf{30.78} & \textbf{14.62} & \textbf{43.05} & \textbf{26.74} & \textbf{35.40} & \textbf{23.85} & \textbf{56.85} & \textbf{37.88}  \\ 
    \bottomrule
    \end{tabular}  
    }
    \end{table*}
 
\section{Spatio-Temoral Prompt Explainability}
\label{app_prompt}
In the main text of our paper, we introduced direction-aware edges and learnable time gates, which will be elaborated in this section. 
The direction-aware edges consist of forward edges $e^t_{forward}$ and backward edges $e^t_{backward}$, the weights of them are trainable.
Specifically, the forward edge point from a region node's previous time step to its current time step, while backward edge point from the current time step back to the previous one, establishing spatio-temporal dependencies between the region node and its past states.
For each time step within the token window, all region nodes share a pair direction-aware edges.
Moreover, all region nodes share a set of the learnable time gating parameter $\gamma$, and the number of time gating parameters is consistent with the size of the token window\{3,7\}. 

Tables~\ref{initialization} presents our spatio-temoral prompt initialization strategy, we set all time gating weights to 1, and the weight of forward edge is initialized to 1, while the weight of backward edge are initialized to 0.5, which conforms to the temporal directionality prior~\cite{yu2017spatio}.
By initializing the trainable parameters as prompts, we aim to guide the pre-trained model to further model spatio-temporal epidemic patterns. 
Meanwhile, final weights of trainable prompted parameters can be used for model explainability.

Figure~\ref{prompt} visualizes the weights of $e^t_{forward}$ and $e^t_{backward}$, as well as the corresponding $\gamma_k$ when token window size is 3.
Moreover, we present a more intuitive set of prompt weight results in Table~\ref{7 day} when the token window size is 7.
Overall, the time gating weights generally show an increasing trend over time despite some fluctuations, demonstrating \name~places greater emphasis on the current timestep and effectively captures the temporal dependencies.
For direction-aware edges, the trainable forward weight always exceed the backward one, adhering to the temporal directionality assumption.

\begin{table}[htbp]
    \centering
    \caption{The initialization strategy of prompt weights in \name.}
    \label{initialization}
    \begin{tabular}{c c c c}
    \toprule
    \textbf{Prompt parameter} & $e^t_{forward}$ & $e^t_{backward}$ & $\gamma_k$ \\
    \midrule
    \textbf{Initialization}  & 1               & 0.5              & 1    \\
    \bottomrule
    \end{tabular}
\end{table}

\begin{table*}[htbp]
\caption{The final weights of prompted weights in EpiLLM.}
\label{7 day}
\resizebox{\textwidth}{!}{
\begin{tabular}{cccccccccc}
\toprule
\textbf{Dataset} & $e^t_{forward}$ & $e^t_{backward}$ & $\gamma_{t-6}$ & $\gamma_{t-5}$ & $\gamma_{t-4}$ & $\gamma_{t-3}$ & $\gamma_{t-2}$ & $\gamma_{t-1}$ & $\gamma_t$ \\ \midrule
\textbf{Italy}   & 0.7964          & 0.6654           & 0.6794       & 0.6664       & 0.6649       & 0.8129       & 0.7898       & 1.0235       & 1.2022     \\
\textbf{Spain}   & 1.2093          & 0.6187           & 0.8695       & 0.8384       & 0.7887       & 0.8225       & 0.8474       & 1.0331       & 1.0455     \\
\textbf{England} & 1.3706          & 0.3904           & 0.9269       & 0.9199       & 0.9325       & 0.9585       & 1.0006       & 1.0581       & 1.1743     \\
\textbf{France}  & 1.0009          & 0.4934           & 1.0056       & 1.0062       & 1.0105       & 1.0109       & 1.0183       & 1.0194       & 1.0265     \\ \bottomrule
\end{tabular}
}
\end{table*}

\section{Case Study}
\label{app_casestudy}
\begin{figure}[htbp]
\centering
\includegraphics[width=0.98\linewidth]{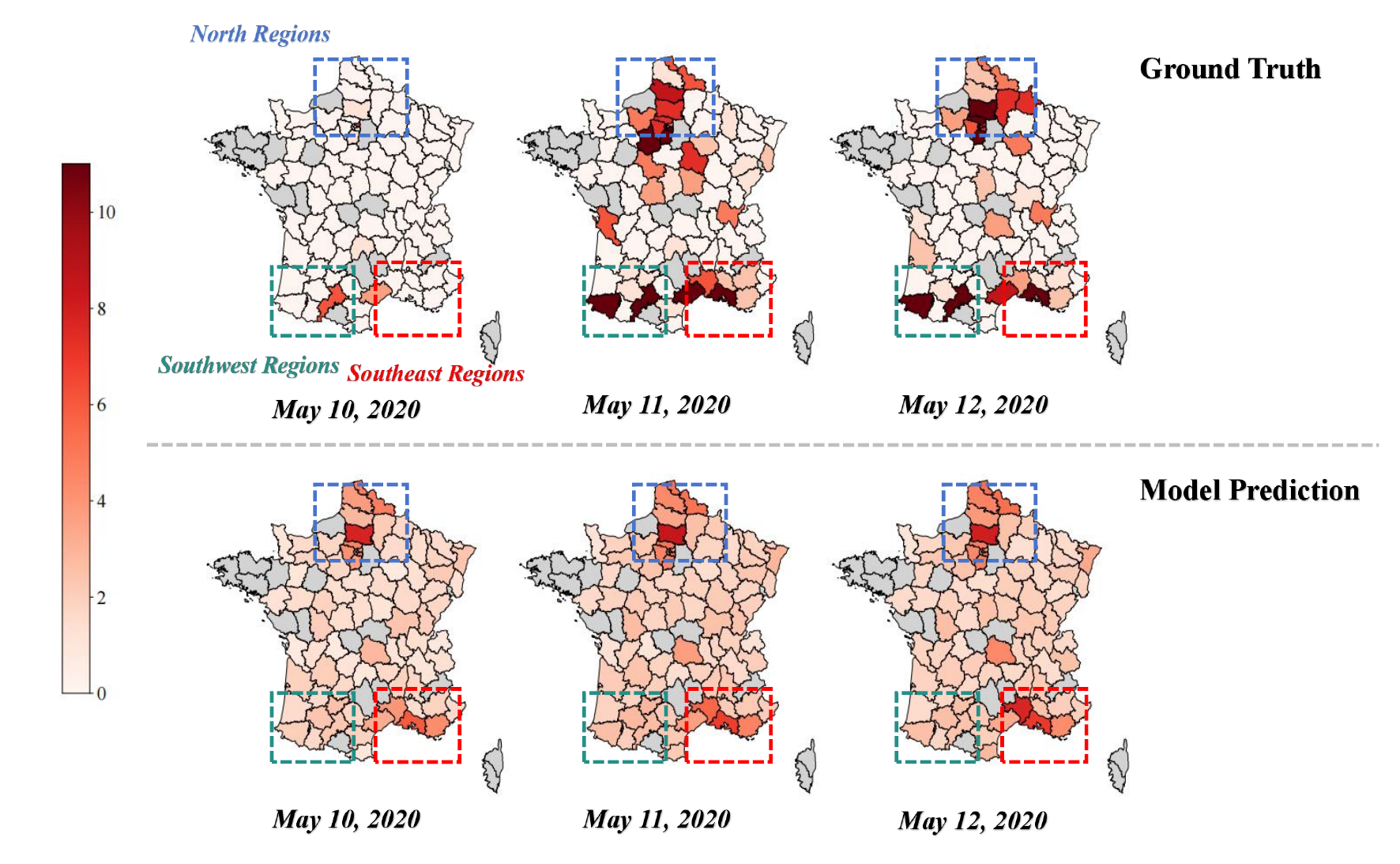}
\caption{Case study of France (part regions) COVID-19 progression during May 10-12, 2020. Areas shaded in gray denote regions with unavailable surveillance records.}
\label{casestudy}
\end{figure}

\begin{figure}[htbp]
\centering
\includegraphics[width=0.98\linewidth]{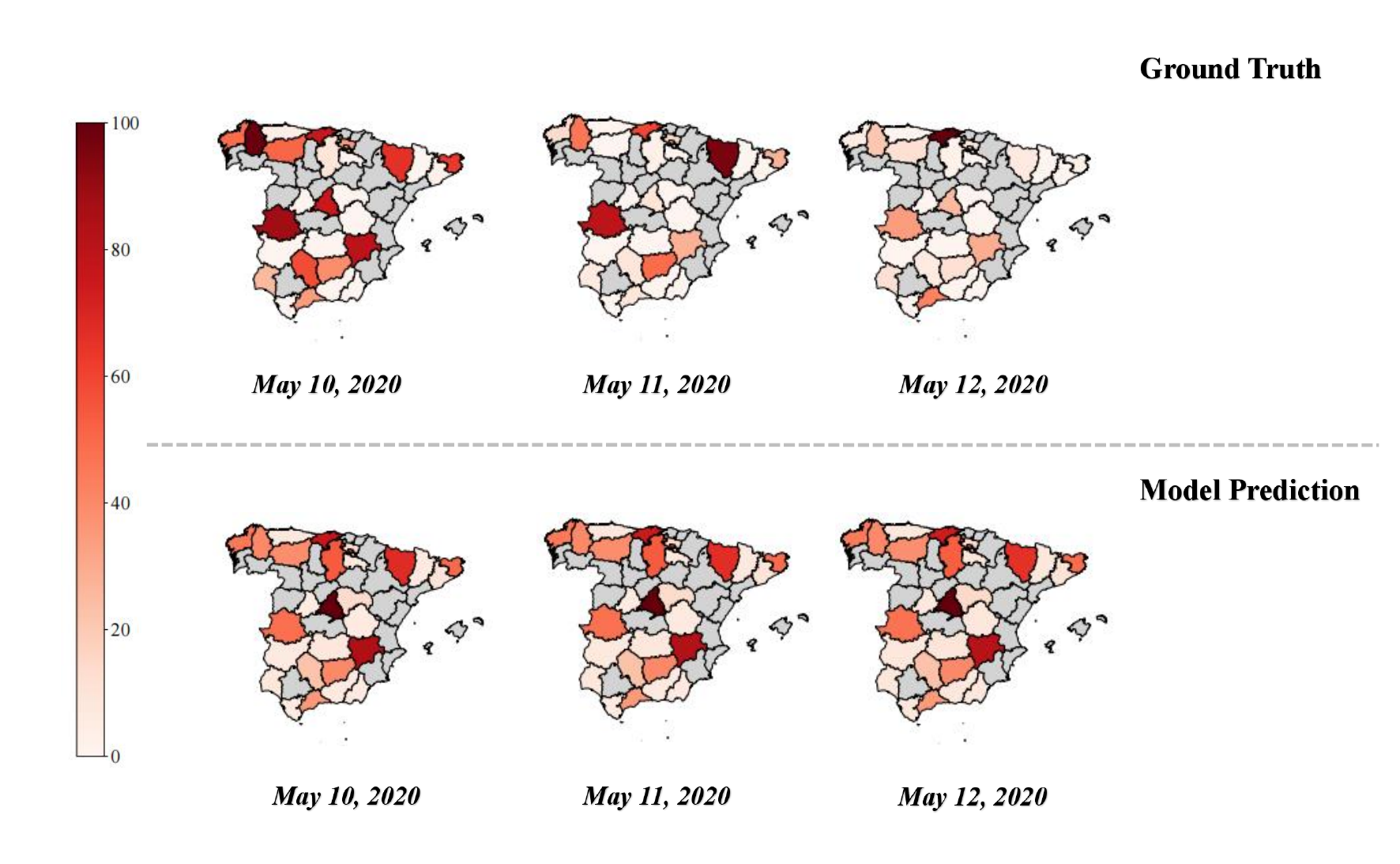}
\caption{Case study of Spain (part regions) COVID-19 progression during May 10-12, 2020. Areas shaded in gray denote regions with unavailable surveillance records.}
\label{casestudy2}
\end{figure}

\begin{figure}[htbp]
\centering
\includegraphics[width=0.98\linewidth]{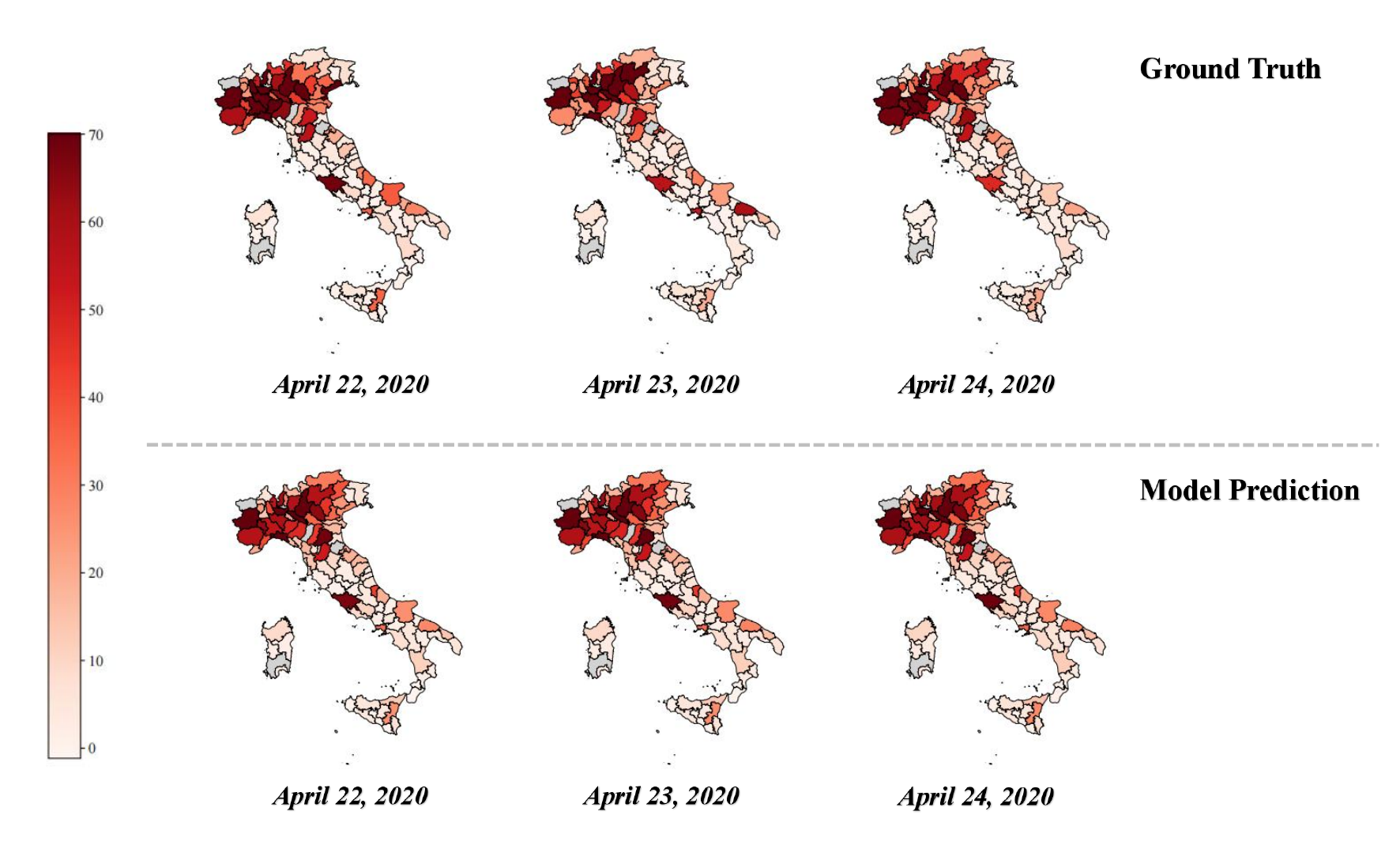}
\caption{Case study of Italy (part regions) COVID-19 progression during April 22-24, 2020. Areas shaded in gray denote regions with unavailable surveillance records.}
\label{casestudy3}
\end{figure}

As can be seen in Figure~\ref{casestudy}, our case analysis features a visualization of the epidemic progression dynamics of COVID-19 in France (part regions) during May 10-12, 2020.
Our case study primarily focuses on analyzing the epidemics across three key regions of France: the north regions, southwest regions, and southeast regions.
The results demonstrate that our proposed model accurately predicted the epidemic progression in both north and southeast regions in France, further validating its effectiveness.
However, discrepancies were observed between the model's predictions and the actual epidemic progression in southwest regions, which can be attributed to the area's sudden outbreak pattern that exceeded the model's real-time response capacity.
Therefore, to address the complex pandemic patterns observed in real-world scenarios, it is imperative to enhance the model's emergency response capability and early-warning capacity, both of which we identify as critical directions for future research.
In addition, we also visualized the epidemic prediction and ground truth for part regions of Spain and Italy in Figure~\ref{casestudy2} and Figure~\ref{casestudy3}.
What we need to emphasize is that in the epidemic forecasting task for regions in Italy, the predictions made by EpiLLM in Figure~\ref{casestudy3} are highly consistent with the actual outcomes.

\end{document}